# Toxic language detection: a systematic review of Arabic datasets

Imene Bensalem[1, 2], Paolo Rosso[3], Hanane Zitouni[4]

[1] ESCF de Constantine, ibensalem@escf-constantine.dz
[2] MISC Lab, Constantine 2 University
[3] Universitat Politècnica de València, prosso@dsic.upv.es
[4] Constantine 2 University, hanane.zitouni@univ-constantine2.dz

**Abstract**

The detection of toxic language in the Arabic language has emerged as an active area of research in recent years, and reviewing the existing datasets employed for training the developed solutions has become a pressing need. This paper offers a comprehensive survey of Arabic datasets focused on online toxic language. We systematically gathered a total of 54 available datasets and their corresponding papers and conducted a thorough analysis, considering 18 criteria across four primary dimensions: availability details, content, annotation process, and reusability. This analysis enabled us to identify existing gaps and make recommendations for future research works. For the convenience of the research community, the list of the analysed datasets is maintained in a GitHub repository[1].

## 1    Introduction

While social media have been and continue to be a force of change in many Arab countries by providing their users with a platform for free expression (ASDA'A BCW, 2021), they are also a breeding ground for spreading toxic discourse, such as hate speech and offensive language.

Social media platforms rely on internal policies, artificial intelligence algorithms, user reporting, and content moderators to fight against inappropriate content (Laub, 2019). Nonetheless, some platforms have been accused of doing little in dealing with content in languages other than English[2]. Moreover, their algorithms have been criticized for occasionally failing to distinguish between toxic language and legitimate criticism, leading to the erroneous removal of accounts or posts belonging to journalists and human rights defenders[3].

---

[1] https://github.com/Imene1/Arabic-toxic-language
[2] See examples on the Arab world in the press article "Social media platforms doing little to combat online hate speech in the Arab world: Experts". https://www.arabnews.com/node/1944116/media and another example about India in "Facebook Says It's Removing More Hate Speech than Ever Before. But There's a Catch", https://time.com/5739688/facebook-hate-speech-languages and "How Facebook neglected the rest of the world, fueling hate speech and violence in India" https://www.washingtonpost.com/technology/2021/10/24/india-facebook-misinformation-hate-speech
[3] See examples of the wrong removal of posts on Gaza and Syria wars in "Facebook's language gaps weaken screening of hate, terrorism", https://www.arabnews.com/node/1954911/media

These weaknesses in the automated methods may lead not only to spreading inappropriate content but also to undermining the freedom of speech, discarding objective discourse, and causing harm to individuals and communities (Gelber & McNamara, 2016), hence the need to fight it.

Toxic language detection has been an active area of research in the last years and recently, many works focused on the Arabic text (Husain & Uzuner, 2021) accompanied by an increasing number of created datasets. However, beyond the widely recognized datasets, whether originating from pioneering efforts (Alakrot et al., 2018; Albadi et al., 2018; Haddad et al., 2019; Mulki et al., 2019; Ousidhoum et al., 2019) or released in the context of shared tasks (Mubarak, Al-Khalifa, et al., 2022; Mubarak, Darwish, et al., 2020; Mulki & Ghanem, 2021a; Zampieri et al., 2020), little is known about other datasets regarding their accessibility, annotation quality, the context in which they were created, and their potential for reuse.

While there are several surveys on toxic language detection (Arora et al., 2023; Balayn et al., 2021; Chhabra & Kumar, 2023; Fortuna & Nunes, 2019; Poletto et al., 2021; Vidgen & Derczynski, 2020), with some of them focusing on Arabic language (Al-Hassan & Al-Dossari, 2019; ALBayari et al., 2021; Alsunaidi et al., 2023; Elzayady et al., 2022; Husain & Uzuner, 2021; Khairy et al., 2021), no initiative has yet focused on the Arabic datasets. Previous seminal initiatives focusing on reviewing resources (Poletto et al., 2021; Vidgen & Derczynski, 2020) are not language-specific and reference only a limited number of Arabic datasets. In addition, due to the rapid growth of this research field, several recent datasets were not included in prior reviews. This highlights the need for additional efforts to review recent works and ensure the visibility of all accessible datasets.

Given the pivotal role of data as fuel for artificial intelligence algorithms (Geiger et al., 2021), and the lack of knowledge about the available datasets on Arabic toxic language detection, this paper aims to systematically and comprehensively collect and review existing datasets dealing with Arabic toxic language. More precisely, our objective is to answer the following research questions.

RQ1- What are the available Arabic datasets dealing with toxic language?

RQ2- What are the specific tasks for which those datasets have been created?

RQ3- How were these datasets annotated, and were the best annotation practices applied?

RQ4- Can these datasets be easily reused for future research?

Taking inspiration from the definition of online toxicity provided by (De Smedt et al., 2021), this systematic review employs the term "toxic language" as a broad concept encompassing various forms of language that have the potential to harm individuals, groups of individuals or entities. It encompasses a wide range of manifestations, including profanity, obscenity, abusiveness, verbal aggression, hate speech and violent language. Those concepts can be further classified into numerous fine-grained categories, such as racism, sexism, and threats. In addition, a text could be considered toxic due to its harmful content, even if it does not involve explicit offensiveness, such as disinformation. Based on this broad definition of toxicity and using a systematic methodology, we aim to achieve comprehensiveness and rigour.

The remainder of this paper is structured as follows: Section 2 presents a review of related works. Section 3 outlines the methodology used to undertake this survey. Section 4 presents an in-depth analysis of the collected datasets. Section 5 discusses the findings and provides recommendations. Finally, Section 6 concludes the paper.

Table 1. Summary of some recent related works.

| Reference | Covered languages | Analysed datasets' aspects | | | #Referenced available Arabic datasets on toxic language | Systematic review? (Methodology) |
|---|---|---|---|---|---|---|
| | | Content | Accessibility | Annotation process | | |
| General surveys on toxic language | | | | | | |
| (Chhabra & Kumar, 2023) | Multiple | ✓ | ✗ | ✗ | 4 | No |
| (Jahan & Oussalah, 2023) | | ✓ | ✓ | ✗ | 6 | Yes (PRISMA)[†] |
| (Mahmud et al., 2023) | | ✗ | ✗ | ✗ | N/A | No |
| (Al-Hassan & Al-Dossari, 2019) | Arabic | ✗ | ✗ | ✗ | 4 | No |
| (Husain & Uzuner, 2021) | | ✓ | ✓ | ✗ | 9 | Yes (N/A) |
| (ALBayari et al., 2021) | | ✓ | ✓ | ✗ | 3 | Yes (N/A) |
| (Khairy et al., 2021) | | ✗ | ✗ | ✗ | N/A | No |
| (Elzayady et al., 2022) | | ✓ | ✗ | ✗ | 10 | No |
| (Alsunaidi et al., 2023) | | ✗ | ✗ | ✗ | 2 | No |
| Surveys of Arabic available resources | | | | | | |
| (Ahmed et al., 2022) | Arabic | ✓ | ✓ | ✗ | 2 | Yes (PRISMA) |
| (Altaher et al., 2022; Alyafeai et al., 2022) | | ✓ | ✓ | ✗ | 11 | Yes (Rowley)[‡] |
| Dataset-oriented surveys on toxic language | | | | | | |
| (Vidgen & Derczynski, 2020) | Multiple | ✓ | ✓ | ✓ | 6 | Yes (PRISMA) |
| (Poletto et al., 2021) | | ✓ | ✓ | ✓ | 6 | Yes (Kitchenham)[§] |
| Our survey | Arabic | ✓ | ✓ | ✓ | **54** | Yes (Kitchenham) |

[†]: PRISMA is a methodology introduced in (Moher et al., 2010).
[‡]: We used the name of its first author to refer to the methodology descried in (Rowley & Slack, 2004).
[§]: We used the name of its first author to refer to the methodology descried in (Kitchenham & Charters, 2007).

## 2 Related works

In recent years, several survey papers have been published on toxic language detection. While some of them cover multiple languages, others focus on Arabic. On the other hand, the majority of them focus on reviewing the approaches, including the classification methods and the features, while few concentrate on the datasets. Below, we provide a summary of the surveys that we deem most relevant to our work.

In their seminal work, Vidgen and Derczynski (2020) analysed more than 60 abusive language datasets including six in Arabic. In their datasets collection, Arabic is the second language after English in terms of the available abusive language datasets. Datasets are analysed in terms of the following aspects: Motivation of the creation, Task definition, Content, and Annotation. The paper provides insightful recommendations on the best practices aiming to build good-quality datasets while avoiding biases and responding to real-world requirements. The authors maintain an online catalogue of abusive language data[4].

---

[4] "Catalog of abusive language data." PLoS 2020. Currently available at Hate Speech Data https://hatespeechdata.com/

Poletto et al. (2021) conducted a systematic review of the available resources on hate speech detection. The analysis of resources, comprising datasets and lexicons, took into account various aspects such as the development methodology, the topical biases, and the language coverage. In this review, six Arabic datasets were referenced.

Husain and Uzuner (2021) focused on offensive language detection works dealing with Arabic. The paper provides the most detailed review of the techniques, resources, and approaches applied in the area. The paper cited 9 publically available datasets published until 2020. However, apart from discussing the inter-annotator agreement of the cited datasets, the paper did not delve into the other details of the annotation process.

In addition to the surveys focusing on toxic language, Masader Plus (Altaher et al., 2022), an online catalogue for Arabic text and speech data resources[5], indexes 11 toxic language datasets among more than 500 Arabic resources[6]. Each dataset is described using 25 attributes (Alyafeai et al., 2022). Besides, Ahmed et al. (2022) provide a scoping survey of the freely available Arabic corpora, but it lists only 2 datasets related to toxic language.

In Table 1, we included additional surveys for comparison purposes. In this table, we listed among the surveys that cover multiple languages only the most recent ones. However, we tried to include an exhaustive list of the surveys focusing on Arabic. As can be observed in the table, there is an absence of systematic reviews focusing on Arabic toxic language datasets. Existing surveys concentrating on datasets are not language-specific, and those focusing on Arabic are not datasets-specific. In addition, most of the surveys that examined Arabic toxic language datasets failed to consider the dataset's accessibility as a criterion for selection, and concentrated on describing the content of the datasets, such as size and labels, while offering limited or no details regarding the annotation process. Furthermore, previous initiatives cataloguing Arabic resources have often overlooked many toxic language datasets, possibly due to the use of topic-independent keywords in their resource search. On top of that, none of the existing works covered more than 11 Arabic toxic language datasets.

Our survey aims to bridge this gap by providing a comprehensive and systematic examination of the currently available Arabic datasets. To ensure comprehensiveness, we strived to employ a diverse range of keywords and conducted our research not only on literature databases but also on datasets' repositories, as elaborated in the following section.

## 3 Methodology

We derived the main phases of undertaking this survey from the guidelines of (Kitchenham & Charters, 2007) for performing systematic literature reviews in software engineering. Since our survey concerns natural language processing, and it requires searching and screening not only literature but also datasets, we adapted this methodology to suit our task. The following subsections provide details on each stage of the methodology, which are search strategy, study selection and quality assessment, data extraction, and data synthesis.

### 3.1 Search strategy

Our search strategy, aiming to collect papers and datasets is based on three stages, which are keywords curation, collecting papers, and collecting datasets' URLs. We executed this search process at different intervals, with the most recent search conducted in October 2023.

---

[5] "Masader Plus. Arabic NLP data catalogue". Available at : https://arbml.github.io/masader/
[6] We have counted the 11 datasets, on Masader Plus, related to toxic language through a manual search using the keywords "Hate speech detection", "Abusive language detection", "Offensive language detection", and "Adult language detection" during our last access on December 21, 2023.

### 3.1.1 Curating keywords

Given our use of *toxic language* as an umbrella term encompassing various related concepts, we have strived to amass a wide array of associated terms sourced from surveys and taxonomies (Banko et al., 2020; Fortuna & Nunes, 2019; Jahan & Oussalah, 2023). The following keywords constitute our compiled list: *Hate speech*, *offensive language*, *offence*, *dangerous language*, *toxic language*, *toxicity*, *abusive language*, *insult*, *harmful language*, *harmfulness*, *cyberbullying*, *racism*, *radicalism*, *misogyny*, *sexism*, *violence*, *hostility*, *profanity*, *sexual harassment*, and *fanaticism*.

Additionally, as the datasets for *stance detection*, *propaganda detection*, *polarised speech*, *polarization*[7] and *social bots detection* may potentially include harmful language, we incorporated these terms into our list of keywords.

For better precision of the results, each of these keywords has been combined, using the operator AND, with two additional keywords, namely *Arabic* and *dataset,* given our focus is on Arabic datasets. Moreover, when the search results are excessively cluttered with irrelevant papers, we occasionally integrate the term *detection* into the search query, particularly if it is not already a part of the employed keyword.

### 3.1.2 Collecting papers

We conducted the research through Google Scholar, focusing on scholarly works, whether published or preprints, written in English and appearing, based on their titles and abstracts, to be related to the detection of online toxic language. If it became evident that a paper did not involve computational work or was focused on social media analysis without considering toxicity or harmfulness, we chose not to download it. For each search query, we reviewed on average around 4 pages of the search results. Beyond that point, the results become generally less relevant.

We examined the downloaded papers with the aim of retaining only those that introduce a new dataset. Therefore, papers describing experiments employing previously published datasets are discarded. We discarded also the papers mentioning clearly that the employed dataset is private.

### 3.1.3 Collecting datasets

After confirming that the paper introduces a new dataset, our goal is to find the dataset's URL. This involves two main phases: (i) scrutinizing the paper's content seeking for URLs and (ii) conducting a web search for the dataset. In the first phase, we established a stepwise process for manually inspecting the paper. If the dataset is not found in one step, we proceed to the next. The steps are as follows:

a. Searching for a data availability statement in the paper.
b. Using "http" as a keyword to identify and then inspect all URLs present in the paper.
c. Scanning the paper's footnotes for any hyperlinks or hints about the dataset's availability.
d. Looking for the term "available" within the paper, as statements about the availability of the dataset typically contain this word.

If the examination of the paper fails to reveal the dataset's URL, we move to the second phase, which is conducting a web search. However, this process is undertaken exclusively in two situations: when the paper explicitly states the current or future availability of the dataset, or when the introduced dataset stands out due to its distinctive task, specific annotation labels, or a notable citation count of its associated paper. In such cases, we formulate our search query using key details like the dataset's name (if available), the paper title, and/or the authors' names. These queries are then submitted to the Google search engine, as well as specific repositories, namely Github, Kaggle, Huggingface, Papers with Code, Open Science Framework, and Zenodo.

---

[7] We included the various morphological variations of the term if any.

Table 2. Inclusion and exclusion criteria

| | |
|---|---|
| **C1** | We exclude a dataset if we do not find a paper describing it. |
| **C2** | We exclude the papers that describe datasets that are unavailable online. |
| **C3** | We exclude papers mentioning the availability of the dataset upon request without providing a dedicated link to make such a request. |
| **C3** | We exclude the datasets on related tasks that may involve toxic content (e.g., disinformation detection, fact-checking, stance detection, sentiment/emotion analysis, sarcasm and irony) but the toxicity is not explicitly annotated. Similarly, we include datasets on other tasks only if the toxic examples are explicitly annotated. For example, although disinformation is likely to be harmful, we include a dataset of this task in our survey only if the content is annotated with labels indicating explicitly the presence of harmfulness. |
| **C5** | We exclude the datasets of tasks that do not involve natural language processing, e.g., datasets containing exclusively network metadata on social media accounts (such as number of followers, number of retweets, etc) without reference to the content. |
| **C6** | We exclude datasets that are merely built by combining existing datasets without additional processing such as re-annotation, filtering…etc, such as the dataset used in (Saketh Aluru et al., 2020). |
| **C7** | We exclusively include annotated datasets, whether the annotations were employed for labelling, tagging, or structuring the text. For example, we did not include several unannotated datasets collected from extremists forums offered by the Dark Web project[8]. |

If, despite these efforts, the dataset's URL cannot be located, we opt to exclude the corresponding paper from our survey. This is for example the case of the papers (Guellil et al., 2022) and (Gharbi et al., 2021) that introduce datasets on the detection of sexism and Tunisian hate speech, respectively. However, they are currently not available.

Note that the aforementioned search phases are overlapping and iterative. For instance, searching for the URL of a dataset may, sometimes, lead to the discovery of other datasets. In such cases, we endeavour to find the papers describing these additional datasets. Additionally, the list of keywords used to search for papers has been gradually extended based on the examined papers.

Finally, it is worth noting that we obtained a few works (papers or datasets) outside of this systematic search. Further details on this point are provided in Appendix A.

---

[8] "Dark Web Forums". Available at Intelligence and Security Informatics Datasets. AZSecure-data.org. https://www.azsecure-data.org/dark-web-forums.html.

Table 3. Criteria used to analyse the collected works.

| Axes | Criteria | Criteria description | Source of the extracted data |
|---|---|---|---|
| Availability details | Publication year | The year when the first paper about the dataset (preprint or published) is made available online. | Paper |
| | Host | The repository or website hosting the dataset. | Dataset |
| Content | Sources | The origin of the texts included in the dataset. | Paper |
| | Size | The total number of examples in the dataset. | Dataset |
| | Task | The intended goal for which the dataset is created, along with the specific addressed task. | Dataset, Paper |
| | Dimension | The number of annotation attributes related to toxicity. | Dataset |
| | Attributes granularity | The number of labels in the toxicity attributes. | Dataset |
| | Dialects and scripts | Specification of whether the text is in modern standard Arabic or dialectal Arabic and whether it is written in the Arabic script or the Latin script (i.e., Arabizi). | Paper |
| Annotation process | Annotation method | Details on how the dataset is annotated. | Paper |
| | Annotators dialect proficiency | Whether the paper discussed the alignment between the dialects mastered by the annotators and those present in the dataset. | Paper |
| | Quality and validation | Measures taken to ensure or check the quality of the dataset annotation. | Paper |
| | Guideline availability | Whether the annotation guideline is made available and the level of detail in its content. | Dataset, Paper |
| | Inter-annotator agreement | Whether the inter-annotator agreement is calculated and how. | Paper |
| | Annotators' support | Whether the annotators receive emotional and financial support. | Paper |
| Reusability | Data-paper consistency | Verification of whether the dataset used in the experiments reported in the paper corresponds to the actual dataset. | Dataset, Paper |
| | Training-test split | Whether the dataset is divided into training and test subsets. | Dataset |
| | Textual content availability | Whether the dataset contains the text of the posts or only their IDs. | Dataset |
| | Raw annotations | Whether the dataset comprises the annotators' raw labels used to form the gold label. | Dataset |

## 3.2 Study selection and quality assessment

Table 2 shows the inclusion and exclusion criteria that we applied to select the papers and the datasets to keep for this systematic review. Note that some of those criteria have been already mentioned in the above subsections.

It is essential to highlight that we opted not to introduce supplementary inclusion and exclusion criteria related to the quality of the works. This decision stems from the fact that our research question integrally investigates quality considerations about datasets, particularly regarding their annotation and reusability, and how information regarding these aspects is reported in the corresponding papers.

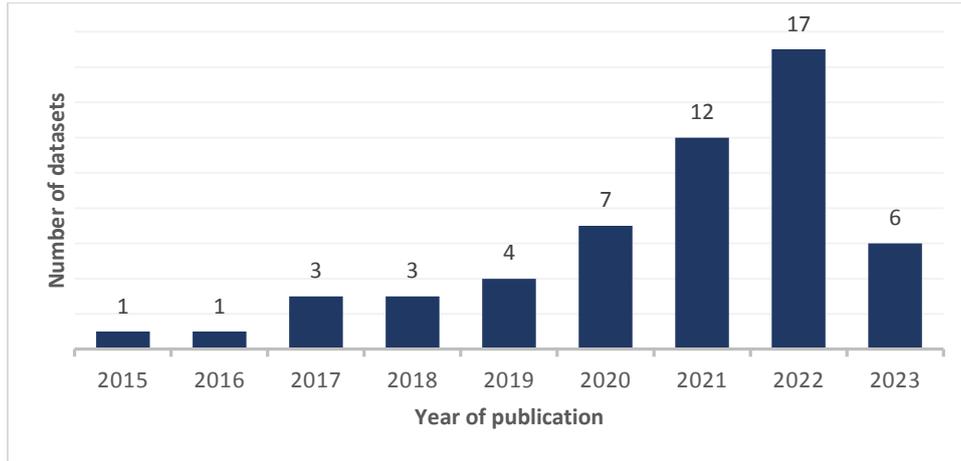

Figure 1. Publication year of the papers corresponding to the toxic language detection datasets included in our collection.

### 3.3  Data extraction

We assigned unique identifiers to the collected datasets and catalogued them, alongside numerous details, in an extensive table. We extracted these details through a thorough examination of each dataset and its corresponding paper, covering various dimensions. The formulation of these dimensions has been influenced by previous surveys, particularly the work of Vidgen and Derczynski (2020), which provided guidelines for best practices in creating and reporting on abusive language detection datasets. However, we have reorganized these dimensions in our unique structure, presenting them into 18 criteria classified under four axes: availability details, content, annotation process, and reusability. Table 3 lists these criteria, defines them, and specifies whether we extracted the details related to each criterion by examining the dataset files or the related paper.

### 3.4  Data synthesis

To attain a comprehensive understanding of the examined datasets and discern trends as well as potential gaps, the final phase in preparing this systematic review is to scrutinize the large table that describes the datasets across the 18 aforementioned criteria. Specifically, for each criterion, we examined thoroughly the details extracted from all works, aiming to cluster them into subsets or identify patterns. The outcomes of these analyses are elaborated in the subsequent section.

## 4  Results

The methodology we employed to collect papers and datasets enabled us to amass 54 works. By 'work,' we refer to a dataset with at least one paper describing it. Note that some datasets are derived from others, with variations in the number of examples or attributes they contain. In such cases, as long as they are shared through separate URLs, referenced in different papers, and exhibit any distinction, even small, we consider them as distinct datasets. An example of this is evident in the datasets initially described in (Alam, Dalvi, et al., 2021) and (Mubarak, Rashed, et al., 2020), which have subsequently been released in various versions utilized in different shared tasks.

To the best of our knowledge, this is the most comprehensive collection of Arabic toxic language datasets published so far. Appendix C lists the references of all the collected datasets along with their URLs. Furthermore, we have established an online catalogue that will be consistently updated (see the link in Endnote 1).

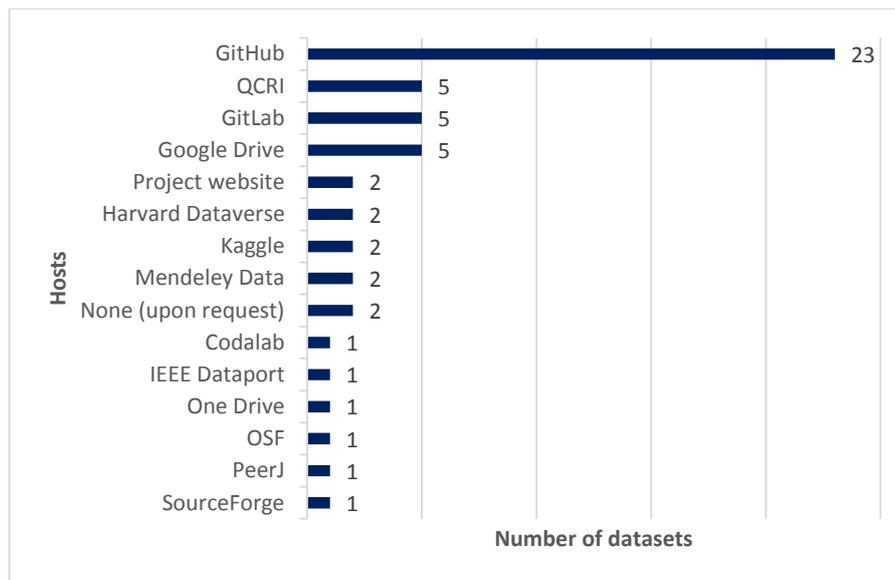

Figure 2. Hosts of toxic language detection datasets.

This section presents our analysis of these datasets based on the 18 criteria described earlier (§3.3), organized into four axes: Availability Details, Content, Annotation process, and Reusability.

## 4.1 Availability details

### 4.1.1 Publication year

Figure 1 illustrates that the initial work, with an accessible dataset, addressing Arabic online toxicity was published in 2015. Subsequently, the number of datasets steadily increased until a significant surge occurred in 2021, with the publication of 12 research papers making their dataset available. Another notable leap took place in 2022, with the release of 17 new datasets. This trend highlights a growing interest in the importance of studying and addressing Arabic online toxicity through the development of dedicated datasets.

### 4.1.2 Host

As depicted in Figure 2, GitHub hosts more than half of the datasets (23 datasets) followed by the Qatar Computational Research Institute (QCRI), which hosts 5 datasets. The rest of the hosts are either public repositories, academic publisher's repositories, or the project's website (the case of the OSCAR datasets), except for 2 datasets that we obtained by requesting the authors. Refer to Appendix C for the full URLs.

## 4.2 Content

### 4.2.1 Source

As can be observed in Table 4, Twitter is the source of text for more than half of the datasets, followed by Facebook and YouTube. Instagram and Reddit were used to build only 2 datasets, while several social media platforms are still underexplored as sources in the Arabic datasets such as Wikipedia and TikTok.

Besides, three datasets are extracted from the web, which are the large-scale corpus OSCAR, where adult content is labelled (Abadji et al., 2022; Jansen et al., 2022), the dataset of the deleted comments from the website of the news channel Aljazeera (Mubarak et al., 2017) and Riabi's et al. (2023) dataset, which includes user comments from a news website in addition to texts from a corpus of song lyrics.

Table 4. Sources of the Arabic datasets on the online toxic language.

| Source | | References |
|---|---|---|
| Social media | Twitter | (Jamal et al., 2015) (Alhelbawy et al., 2016) (Abozinadah & Jones, 2017) (Mubarak et al., 2017) (Albadi et al., 2018) (De Smedt et al., 2018) (Albadi et al., 2019) (Mulki et al., 2019) (Ousidhoum et al., 2019) (Alsafari et al., 2020) (Alshalan & Al-Khalifa, 2020) (Alshehri et al., 2020) (Chowdhury et al., 2020) (Zampieri et al., 2020) (Alam, Dalvi, et al., 2021) (Alam, Shaar, et al., 2021) (Aldera et al., 2021) (Alqmase et al., 2021) (Awane et al., 2021) (Hadj Ameur & Aliane, 2021) (Mubarak et al., 2021) (Mubarak, Rashed, et al., 2020) (Mulki & Ghanem, 2021b) (Mulki & Ghanem, 2021a) (Nakov et al., 2021) (Shaar et al., 2021) (Alam et al., 2022) (Almanea & Poesio, 2022) (Badri et al., 2022) (Fraiwan, 2022) (M. Habash & Daqour, 2022) (Khairy et al., 2022) (Mubarak, Hassan, et al., 2022) (Nakov et al., 2022) (Obeidat et al., 2022) (Shannag et al., 2022) (Alduailaj & Belghith, 2023) (Bensalem et al., 2023) (Mazari & Kheddar, 2023) (Raïdy & Harmanani, 2023) |
| | Facebook | (Haddad et al., 2019) (Abainia, 2020) (Chowdhury et al., 2020) (Awane et al., 2021) (Badri et al., 2022) (Boucherit & Abainia, 2022) (M. Habash & Daqour, 2022) (Khairy et al., 2022) (Bensalem et al., 2023) (Mazari & Kheddar, 2023) |
| | YouTube | (Alakrot et al., 2018) (Haddad et al., 2019) (Chowdhury et al., 2020) (Awane et al., 2021) (Albadi et al., 2022) (Badri et al., 2022) (Mohdeb et al., 2022) (Alduailaj & Belghith, 2023) (Bensalem et al., 2023) (Essefar et al., 2023) (Mazari & Kheddar, 2023) |
| | Instagram | (Al Bayari & Abdallah, 2022) |
| | Reddit | (M. Habash & Daqour, 2022) |
| Web | User comments on news websites | (Mubarak et al., 2017) (Riabi et al., 2023) |
| | Webcrawl | (Abadji et al., 2022) (Jansen et al., 2022) |
| Other | Song lyrics corpus | (Riabi et al., 2023) |
| | Hand-crafted | (Röttger et al., 2022) (Ousidhoum et al., 2021) |

Several datasets are multi-sources (Alduailaj & Belghith, 2023; Awane et al., 2021; Badri et al., 2022; Bensalem et al., 2023; Chowdhury et al., 2020; M. Habash & Daqour, 2022; Haddad et al., 2019; Khairy et al., 2022; Mazari & Kheddar, 2023; Riabi et al., 2023), which explains their presence with more than one source in Table 4.

Some datasets originated from other sources. For example, the dataset of (Ousidhoum et al., 2021) is hand-crafted based on patterns to assess the toxicity generated by large language models (see further details on this dataset in §4.2.3). The dataset known as Multilingual Hatecheck (Röttger et al., 2022) is a functional test that consists of synthetic texts in 10 languages, including Arabic. Within this dataset, the Arabic subset constitutes 3570 cases of both hateful and non-hateful content. These cases were carefully crafted by language experts using 716 templates, where the hate speech target and the slur word vary across the cases. The purpose of this dataset is to assess hate speech detection models based on 25 fine-grained functionalities. Each functionality reflects the ability of the model to classify correctly either a specific kind of hate speech or non-hate speech (e.g., implicit derogation, threat, non-hateful messages including profanity, counter speech), or cases exhibiting lexical or syntactic phenomena (e.g., negation, questions, and spelling variation).

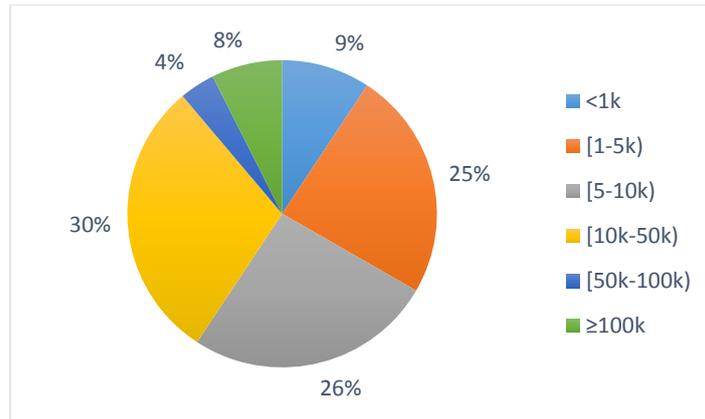

Figure 3. Sizes of the Arabic toxic language detection datasets.

### 4.2.2 Size

As illustrated in Figure 3, most of the datasets have a size between 1k and 50k examples, which is in line with the size of datasets in English and other languages according to a recent survey on hate speech detection (Jahan & Oussalah, 2023). Building larger annotated datasets is a challenge, as it may require recruiting many annotators, which makes the task expensive. It might also be ethically critical if each annotator is required to label a large number of examples, as this would expose them to harmful content for an extended period (Kirk et al., 2022). The largest datasets in our collection are the 2 versions of OSCAR, namely 22.1 (Abadji et al., 2022) and 23.2 (Jansen et al., 2022), which is an immense multilingual textual corpus employed for training large language models. Unlike their predecessors, these recent versions include automatic annotations for adult content, albeit in a very small proportion. Another dataset is about sports fanaticism detection (Alqmase et al., 2021). It contains more than 230k examples, and its training subset has been annotated automatically based on a lexicon of fanaticism terms.

### 4.2.3 Task

The datasets we have collected deal with the toxic language in the context of different tasks. Upon examining those tasks, we classified the datasets into 4 generic categories and 18 subcategories representing the context or the goal for which they were created.

As shown in Table 5, the first category involves the datasets that were built mainly for training and testing toxic language detectors in textual data. Most of the datasets in this category refer to general offensive and abusive language, hate speech, and cyberbullying. We group together these tasks, despite their different names, because based on their definitions provided in the datasets' papers, differentiating them from each other is not straightforward. For example, some papers describe the task as offensive language detection (Alakrot et al., 2018; Mubarak et al., 2017; Zampieri et al., 2020), but their guidelines/papers include the description of hate speech as one of the cases that should be labelled as offensive. In addition to the general toxicity, some datasets address specific kinds of toxicity, such as misogyny, religious hate speech, and violence.

In the second category, datasets are designed for addressing other text processing tasks, but they include labelled toxic texts. The toxicity labels serve either to support the main task or to accommodate multitasking purposes. Below, we provide some examples.

Alam, Dalvi, et al. (2021) created a dataset of tweets aiming to fight disinformation about COVID-19. For each tweet, the annotators have to answer 7 questions revolving around the factuality and the harmfulness of the tweet. The harmfulness was addressed in two questions asking the annotators *whether the tweet is harmful and why*. This dataset has several versions. One of them was used in the NLP4IF-2021 shared tasks on fighting the COVID-19 Infodemic (Shaar et al., 2021). Another version was provided within the harmfulness detection subtask of CLEF-2022 Check That! Lab's task 1 (Nakov et al., 2022), which is about identifying relevant claims in tweets.

Another example is the dataset of (Obeidat et al., 2022) which is labelled with 19 fine-grained classes of misinformation related to the COVID-19 pandemic including the class *Hate and exclusion*. The authors believe that fine-grained annotation is more helpful for fact-checking-worthiness and provides more insight into the harm level of misleading tweets.

Propaganda can use toxic language to achieve its objectives. This is apparent in the definitions of the two propaganda techniques *Appeal to fear/prejudices*[9] and *Smear*[10], which are used among more than 15 techniques to annotate the propaganda datasets described in (Nakov et al., 2021) (Alam et al., 2022).

The dataset of (Jamal et al., 2015) is the only one in our pool that was built within the context of a social sciences study. The goal of its creation is to estimate the proportion of anti-Americanism in Arabic tweets in relation to different social and political events, namely the release of the movie Innocence of Muslims (which was considered offensive to Muslims), the hurricane Sandy and some political events in Syrian, Egypt, and Libya among other events. Although this dataset was not constructed for evaluating language models, its diverse range of topics and categories makes it suitable for training models for various tasks, namely abusive language detection, stance detection, and sentiment analysis.

OSCAR is a very large multilingual dataset crawled from the web and conceived to allow the training of large language models. Due to quality concerns regarding such datasets (Kreutzer et al., 2022), harmful text, particularly adult texts, is annotated in the recent versions of OSCAR (Abadji et al., 2022; Jansen et al., 2022). By having this annotation, researchers and practitioners can apply filtering mechanisms to remove harmful text from the dataset, ensuring that their models are not trained on such content.

The third category of datasets concerns multimodal datasets, which are only two. The work of (Albadi et al., 2022) is about the detection of YouTube videos with religious hate speech. The dataset used to train and evaluate the proposed solution comprises a variety of textual and non-textual features such as the video's title and description, the number of likes, the video duration, etc. Another work in this category is the one of (M. Habash & Daqour, 2022). Although it is a graduation project, this work is particularly interesting as it represents the first attempt to detect Arabic hateful memes. The constructed dataset is composed of 1674 memes and their texts[11] with 2 annotation levels: ternary and fine-grained.

The datasets in the fourth category address the issue of toxicity but do not fall into the aforementioned categories. In this category, Albadi et al. (2019) focused on the detection of hateful bots. They constructed a dataset specifically for this task. The authors found that bots were responsible for almost 11% of hateful tweets, showing that bot detection can be useful in detecting hate speech. Additionally, Ousidhoum et al. (2021) built a dataset to assess the toxicity generated by large language models. It is composed of cloze statements in Arabic, English, and French. Each statement starts with the name of a social group (e.g., African, Muslim, etc.) followed by an everyday action (e.g., cooking, hiking, etc.), and finally, a causal clause where the word representing the reason for the action is masked. Large language models (AraBERT (Antoun et al., 2020) in the case of Arabic) were used to predict the masked word. The experiments revealed that those models tend to generate stereotypical, insulting, and confusing reasons for simple actions.

---

[9] "Seeking to build support for an idea by instilling anxiety and/or panic in the population towards an alternative. In some cases, the support is built based on preconceived judgements" (Alam et al., 2022).
[10] "A smear is an effort to damage or call into question someone's reputation, by propounding negative propaganda. It can be applied to individuals or groups" (Alam et al., 2022).
[11] The dataset's GitHub repository exclusively contains the text extracted from memes but not the image files.

Table 5. Tasks addressed by the Arabic datasets that deal with toxic language. The references in **bold** concern datasets that have been used in shared tasks.

| Task details | References |
|---|---|
| Toxic language detection task in text | |
| General HS/AB/OFF/CB | (Mubarak et al., 2017) (Alakrot et al., 2018) (Haddad et al., 2019) (Mulki et al., 2019) (Ousidhoum et al., 2019) (Alshalan & Al-Khalifa, 2020) **(Zampieri et al., 2020)** (Chowdhury et al., 2020) (Alsafari et al., 2020) (Awane et al., 2021) **(Mubarak, Rashed, et al., 2020)** (Badri et al., 2022) **(Mubarak, Hassan, et al., 2022)** (Boucherit & Abainia, 2022) (Röttger et al., 2022) (Khairy et al., 2022) (Alduailaj & Belghith, 2023) (Bensalem et al., 2023) (Essefar et al., 2023) (Mazari & Kheddar, 2023) |
| Radicalism/Extremism/ Jihadist HS/ Religious HS | (De Smedt et al., 2018) (Albadi et al., 2018) (Aldera et al., 2021) (Fraiwan, 2022) |
| Misogyny | **(Mulki & Ghanem, 2021a)** (Mulki & Ghanem, 2021b) **(Almanea & Poesio, 2022)** |
| Anti-refugees HS | (Mohdeb et al., 2022) |
| Sport fanaticism | (Alqmase et al., 2021) |
| Violence | (Alhelbawy et al., 2016) (Alshehri et al., 2020) (Mubarak, Hassan, et al., 2022) |
| Adult content | (Abozinadah & Jones, 2017) (Mubarak et al., 2021) |
| Toxic texts labelled in the context of other tasks | |
| Misinformation detection | (Alam, Dalvi, et al., 2021) (Alam, Shaar, et al., 2021) **(Hadj Ameur & Aliane, 2021)** (Nakov et al., 2021) **(Shaar et al., 2021)** (Obeidat et al., 2022) **(Nakov et al., 2022)** |
| Propaganda detection | (Nakov et al., 2021) **(Alam et al., 2022)** |
| Sentiment analysis | (Al Bayari & Abdallah, 2022) (Raïdy & Harmanani, 2023) |
| Social science study | (Jamal et al., 2015) |
| Multitask dataset in Arabizi | (Abainia, 2020) |
| Treebank | (Riabi et al., 2023) |
| Large-scale datasets cleaning | (Abadji et al., 2022) (Jansen et al., 2022) |
| Multimodal toxic language detection | |
| Hate speech in videos | (Albadi et al., 2022) |
| Hateful memes detection | (M. Habash & Daqour, 2022) |
| Other tasks dealing with toxicity | |
| Hateful bot detection | (Albadi et al., 2019) |
| Probing the toxicity generated by the large language models | (Ousidhoum et al., 2021) |

It is noteworthy that certain datasets, highlighted in bold in Table 5, have served as benchmarks in 9 shared tasks. Additional details on these shared tasks are provided in Appendix B.

### 4.2.4 Dimension

We use the term dimension to refer to the number of attributes used to describe each example in the dataset. Note that the attributes related to the post, the user profile, the network (e.g., ID, number of followers, number of likes…, etc.) or the raw annotations (§4.4.4) are not considered in this analysis. We have observed two types of datasets: those with one annotation attribute indicating the presence or not of toxicity (e.g., *Offensive*, *Not offensive*) or its type (e.g., *Abusive*, *Hate*, *Normal*), and those with two or more annotation attributes. In the latter case, the annotation attributes typically represent the following:

**Distinct tasks** For example, the dataset of (Abainia, 2020) is annotated for several tasks, namely gender identification, abusive language detection, code-switching detection and emotion analysis.

**Aspects of toxicity** We mean by aspect any attribute providing further details on the main task. Typically, the aspects represent the toxicity fine-grained types or targets. To illustrate, the dataset of (Mulki & Ghanem, 2021b) encompasses an attribute on misogyny identification and two additional attributes determining two aspects of that misogyny: the target (*active*, *passive*) and the fine-grained type (*discredit*, *dominance*, *stereotyping*, etc.).

We observed that the target annotation in our datasets collection is either coarse-grained, determining whether or not the target is mentioned explicitly (Mulki & Ghanem, 2021b; Ousidhoum et al., 2019), or fine-grained determining the attacked group of people (e.g., Muslims, Jews, immigrants, etc.) (Albadi et al., 2022; Alsafari et al., 2020; Ousidhoum et al., 2019; Röttger et al., 2022) or the targeted personal attribute (e.g., race, gender, etc.) (M. Habash & Daqour, 2022; Mubarak, Hassan, et al., 2022; Shannag et al., 2022).

Finally, the datasets of (Shannag et al., 2022) and (Mazari & Kheddar, 2023) are the only ones incorporating an attribute that determines the domain. In (Shannag et al., 2022) for example, this attribute includes the labels: *news*, *celebrities*, *gaming* and *sport*.

**Various toxicity concepts** For instance, in (Mubarak, Hassan, et al., 2022), the examples are labelled as offensive language or not, at the first attribute. If an example is offensive, it is further categorized as hate speech, vulgar language, or/and violent speech, at 3 additional binary attributes.

**Information on the language, the dialect, or the script** For example, in the dataset described in (Mohdeb et al., 2022), which is about hate speech against African refugees in Algeria, the examples are labelled as Arabic, Arabizi, French, or English.

As could be observed in Table Table 6, several datasets involve multiple types of attributes, such as the dataset introduced by Alam, Shaar, et al. (2021). We classified this dataset as multi-task since it comprises various tasks, revolving around COVID-19 misinformation and one of these tasks is the detection of social harmfulness. Additionally, this dataset comprises an attribute that specifies the type of harm if any (i.e., an aspect), namely *bad cure*, *xenophobia*, and *panic*, among others.

Table 6. Types of attributes in the Arabic toxic language datasets.

| | Attribute' type | | References |
|---|---|---|---|
| One-dimension datasets | Presence of toxicity or its types | | (Jamal et al., 2015) (Alhelbawy et al., 2016) (Mubarak et al., 2017) (Mubarak et al., 2017) (Abozinadah & Jones, 2017) (De Smedt et al., 2018) (Albadi et al., 2018) (Alakrot et al., 2018) (Albadi et al., 2019) (Haddad et al., 2019) (Mulki et al., 2019) (Alshalan & Al-Khalifa, 2020) (Alshehri et al., 2020) (Zampieri et al., 2020) (Aldera et al., 2021) (Alqmase et al., 2021) (Awane et al., 2021) (Alam et al., 2022) (Mubarak et al., 2021) (Almanea & Poesio, 2022) (Badri et al., 2022) (Fraiwan, 2022) (Khairy et al., 2022) (Nakov et al., 2022) (Alduailaj & Belghith, 2023) (Essefar et al., 2023) |
| Multi-dimension datasets | Aspects of toxicity | Fine-grained types of toxicity sometimes with other discourse' types | (Ousidhoum et al., 2019) (Alam, Dalvi, et al., 2021) (Alam, Shaar, et al., 2021) (Mulki & Ghanem, 2021b) (Mulki & Ghanem, 2021a) (Nakov et al., 2021) (Röttger et al., 2022) |
| | | Coarse-grained target | (Ousidhoum et al., 2019) (Mulki & Ghanem, 2021b) |
| | | Fine-grained target | (Ousidhoum et al., 2019) (Alsafari et al., 2020) (Albadi et al., 2022) (M. Habash & Daqour, 2022) (Mubarak, Hassan, et al., 2022) (Röttger et al., 2022) (Shannag et al., 2022) |
| | | Domain | (Shannag et al., 2022) (Mazari & Kheddar, 2023) |
| | Various toxicity concepts (related through a hierarchy or not) | | (Mubarak, Rashed, et al., 2020) (Alsafari et al., 2020) (Chowdhury et al., 2020) (Mubarak, Hassan, et al., 2022) (Shannag et al., 2022) (Mazari & Kheddar, 2023) |
| | Multi-task datasets | | (Abainia, 2020) (Alam, Dalvi, et al., 2021) (Alam, Shaar, et al., 2021) (Hadj Ameur & Aliane, 2021) (Nakov et al., 2021) (Shaar et al., 2021) (Al Bayari & Abdallah, 2022) (Obeidat et al., 2022) (Raïdy & Harmanani, 2023) (Riabi et al., 2023) |
| | Language or dialectal labels | | (Abainia, 2020) (Hadj Ameur & Aliane, 2021) (Al Bayari & Abdallah, 2022) (Boucherit & Abainia, 2022) (Mohdeb et al., 2022) (Bensalem et al., 2023) |

### 4.2.5 Attributes granularity

We use the term granularity to refer to the number of classes, in one attribute, used to annotate the dataset examples. If the dataset is multi-task, we consider in our analysis only the attributes related to toxicity.

As shown in Table 7, most of the datasets address toxicity detection as a binary task distinguishing between toxic and non-toxic texts. There are also available datasets that enable ternary or multi-class classification approaches, which allow for a more nuanced understanding of the toxic content and can help design targeted interventions or moderation strategies. Typically, the ternary annotation scheme distinguishes between the clean language and two major kinds of toxic language: (i) hate speech, which targets and denigrates specific groups of people, possibly even in a subtle manner, and (ii) abusive language, which may include vulgar, profane, or obscene words but does not show hate and may not be directed towards a specific target. Among the datasets with a ternary annotation scheme, a single dataset stands out due to its different labels (Fraiwan, 2022). This dataset is about radicalism, and its annotation labels are *Radical*, *Religious but not radical*, and *Unrelated*.

Table 7. Granularity of the toxic language attributes.

| | References |
|---|---|
| Binary | (Abozinadah & Jones, 2017) (De Smedt et al., 2018) (Albadi et al., 2018) (Alakrot et al., 2018) (Abainia, 2020) (Alsafari et al., 2020) (Alshehri et al., 2020) (Chowdhury et al., 2020) (Zampieri et al., 2020) (Alam, Shaar, et al., 2021) (Aldera et al., 2021) (Alqmase et al., 2021) (Hadj Ameur & Aliane, 2021) (Mubarak et al., 2021) (Mubarak, Rashed, et al., 2020) (Mulki & Ghanem, 2021b) (Mulki & Ghanem, 2021a) (Shaar et al., 2021) (Almanea & Poesio, 2022) (Khairy et al., 2022) (Mubarak, Hassan, et al., 2022) (Nakov et al., 2022) (Röttger et al., 2022) (Shannag et al., 2022) (Alduailaj & Belghith, 2023) (Bensalem et al., 2023) (Essefar et al., 2023) (Mazari & Kheddar, 2023) (Riabi et al., 2023) |
| Ternary | (Mubarak et al., 2017) (Haddad et al., 2019) (Mulki et al., 2019) (Alshalan & Al-Khalifa, 2020) (Chowdhury et al., 2020) (Alsafari et al., 2020) (Badri et al., 2022) (Boucherit & Abainia, 2022) (Fraiwan, 2022) (Awane et al., 2021) (M. Habash & Daqour, 2022) (Albadi et al., 2022) |
| Multi-class | (Jamal et al., 2015) (Alhelbawy et al., 2016) (Ousidhoum et al., 2019) (Alsafari et al., 2020) (Alam, Dalvi, et al., 2021) (Alam, Shaar, et al., 2021) (Mulki & Ghanem, 2021b) (Mulki & Ghanem, 2021a) (Nakov et al., 2021) (Abadji et al., 2022) (Jansen et al., 2022) (Alam et al., 2022) (Albadi et al., 2022) (Al Bayari & Abdallah, 2022) (M. Habash & Daqour, 2022) (Mohdeb et al., 2022) (Mubarak, Hassan, et al., 2022) (Obeidat et al., 2022) (Röttger et al., 2022) (Shannag et al., 2022) (Raïdy & Harmanani, 2023) |
| Score | (Albadi et al., 2019) |

There are datasets where the set of labels include multiple classes, addressing the concerned task as a multi-class classification problem. Those classes represent typically fine-grained categories of toxic language or various kinds of text with toxic language being represented by one or more classes. A special case in our pool is represented by the datasets described in (Alam, Dalvi, et al., 2021; Alam, Shaar, et al., 2021). In those datasets, the classes represent a Likert scale showing the confidence level of the annotator regarding whether the text is harmful to society or not. The scale comprises 5 classes ranging from *No, definitely not harmful* to *Yes, definitely harmful*. For illustrations, refer to the examples provided in Table 8. Note that some datasets comprise more than one toxicity attribute with different granularities, such as (Alsafari et al., 2020), which is a dataset that has a hierarchical annotation scheme with 3 levels: binary, ternary and multi-classes.

Finally, in (Albadi et al., 2019) dataset, user accounts that disseminate hate speech were labelled by experts, assigning them a score ranging from 0 to 5, which indicates their similarity to either human or bot behaviour.

Table 8. Examples of multi-class attributes. In the column Labels, the classes indicating the presence of toxicity are represented in bold.

| How the toxic language is represented? | Reference | Attribute description | Labels (toxic labels in bold) |
|---|---|---|---|
| Toxic language among other clean discourses | (Raïdy & Harmanani, 2023) | Type of discourses | (**sectarianism, sexism, racism, bullying, foul language**, sarcasm, joke, courtesy words, saying, known fact, None) |
| Fine-grained categories of toxic language | (Mubarak, Hassan, et al., 2022) | Hate speech categories | (**race/ethnicity/nationality, religion/belief, ideology, disability/disease, social class, gender**, not hate speech) |
| Likert scale | (Alam, Shaar, et al., 2021) (Alam, Dalvi, et al., 2021) | To what extent does the tweet appear to be harmful to society, person(s), company(s) or product(s)? | (No. Definitely not harmful, No. Probably not harmful, Not sure, **Yes. probably harmful, Yes. definitely harmful**) |

### 4.2.6 Dialects and scripts

The Arabic language is known for its rich diversity of dialects, with vernacular variations that span across different regions and countries. On the other hand, Modern Standard Arabic (MSA), which is the formal form of Arabic used in education, news broadcasting, and formal writing, is limited to specific contexts (N. Habash, 2010). Our datasets collection reflects this reality. As shown in Table 9, dialectal Arabic predominates, occurring in the vast majority of the datasets. On the other hand, only two datasets are entirely in MSA: one is the propaganda detection dataset sourced from the Twitter pages of news agencies (Alam et al., 2022), while the other involves synthetic textual patterns serving to investigate the toxicity generated by language models (Ousidhoum et al., 2021).

Some datasets purposefully comprise texts in one dialect, whereas others are created to encompass multiple dialects[12]. The multi-dialect datasets constitute the majority, with 37 datasets. Following are the datasets in the Algerian dialect, which are five.

While texts in Latin letters are usually discarded when building Arabic datasets, we have found in our collection datasets that comprise Arabizi (Alakrot et al., 2018; Boucherit & Abainia, 2022; Mazari & Kheddar, 2023; Mohdeb et al., 2022; Röttger et al., 2022) or are entirely written in this script (Abainia, 2020; Bensalem et al., 2023; Raïdy & Harmanani, 2023). Arabizi is a written form of informal Arabic that uses only Latin letters and numbers (see the references labelled with the superscript [rbz] in Table 9 ). As can be observed in the table, most of the datasets comprising Arabizi are in the Algerian dialect.

Besides, some datasets are multilingual (highlighted in Bold in Table 9). Nonetheless, the examples of each language are saved in a separate file, except the dataset of De Smedt et al. (2018), which comprises several languages in the same file.

### 4.3 Annotation process

The annotation of toxic language plays a crucial role in developing accurate and reliable toxic language detection systems. Providing information on the annotation method, the number of annotators per comment, the quality validation approach, the adopted guidelines and the resolution of annotation disagreements, all contribute to enhancing the transparency of the annotation process. This transparency allows leveraging the dataset while being aware of its potential limitations or biases. The following subsections elaborate on how these aspects were managed in the Arabic toxic language works within our datasets pool.

---

[12] All the datasets that are not entirely in MSA and did not target a specific dialect were considered mixed, meaning they may contain texts in MSA and others in various dialects.

Table 9. Dialects present in the Arabic toxic language datasets. The references in **bold** concern multi-language datasets. The superscript ʳᵇᶻ indicates that the dataset is partially or entirely in Arabizi.

| Dialects | | | References |
|---|---|---|---|
| Only MSA | | | **(Ousidhoum et al., 2021) (Alam et al., 2022)** |
| Mono-dialect | Maghrebi | Algerian | (Abainia, 2020)ʳᵇᶻ (Mohdeb et al., 2022)ʳᵇᶻ (Boucherit & Abainia, 2022)ʳᵇᶻ (Mazari & Kheddar, 2023)ʳᵇᶻ (Riabi et al., 2023)ʳᵇᶻ |
| | | Moroccan | (Essefar et al., 2023) |
| | | Tunisian | (Haddad et al., 2019) |
| | Egyptian | Egyptian | (Mubarak et al., 2017, Twitter dataset) **(Röttger et al., 2022)**ʳᵇᶻ |
| | Levantine | Lebanese | (Mulki et al., 2019) (Mulki & Ghanem, 2021b) (Raïdy & Harmanani, 2023)ʳᵇᶻ |
| | Gulf | Middle Gulf | (Alsafari et al., 2020) |
| | | Saudi | (Alshalan & Al-Khalifa, 2020) (Alqmase et al., 2021) |
| Mixed (various dialects and potentially MSA) | | | (Jamal et al., 2015) (Alhelbawy et al., 2016) (Abozinadah & Jones, 2017) (Mubarak et al., 2017, Aljazeera dataset) (Alakrot et al., 2018)ʳᵇᶻ (Albadi et al., 2018) **(De Smedt et al., 2018)** (Albadi et al., 2019) **(Ousidhoum et al., 2019)** (Alshehri et al., 2020) (Chowdhury et al., 2020) **(Zampieri et al., 2020) (Alam, Dalvi, et al., 2021) (Alam, Shaar, et al., 2021)** (Aldera et al., 2021) (Awane et al., 2021) (Hadj Ameur & Aliane, 2021) (Mubarak et al., 2021) (Mubarak, Rashed, et al., 2020) (Mulki & Ghanem, 2021a) **(Nakov et al., 2021) (Shaar et al., 2021) (Abadji et al., 2022) (Jansen et al., 2022)** (Albadi et al., 2022) (Al Bayari & Abdallah, 2022) (Almanea & Poesio, 2022) (Badri et al., 2022) (Fraiwan, 2022) (M. Habash & Daqour, 2022) (Khairy et al., 2022) (Mubarak, Hassan, et al., 2022) **(Nakov et al., 2022)** (Obeidat et al., 2022) (Shannag et al., 2022) (Alduailaj & Belghith, 2023) (Bensalem et al., 2023)ʳᵇᶻ |

### 4.3.1 Annotation method

Toxic language annotation can be performed manually by human annotators, automatically based on predefined rules or trained models, or can be derived from prior annotations or actions taken by platform moderators. The three former methods were used in the examined datasets.

As depicted in Table 10, most of the datasets underwent manual annotation. Within this category, some datasets were annotated through crowdsourcing platforms, while others relied on a small number of annotators selected based on criteria such as expertise in the annotation task (Röttger et al., 2022), domain knowledge (Fraiwan, 2022; M. Habash & Daqour, 2022; Riabi et al., 2023), age (Al Bayari & Abdallah, 2022), beliefs and gender (Almanea & Poesio, 2022) and the nationality (Alakrot et al., 2018).

In certain works, such as (Aldera et al., 2021; Hadj Ameur & Aliane, 2021), annotators are identified as experts without specifying their domain of expertise. For many datasets, annotators were selected simply because they are native Arabic speakers (sometimes the fluency in a specific dialect is mentioned (§4.2.6)) with a higher education level, such as (Abainia, 2020; Alsafari et al., 2020; Boucherit & Abainia, 2022; Mubarak et al., 2021; Shannag et al., 2022). Furthermore, some datasets were annotated by their respective authors (Alshehri et al., 2020; Badri et al., 2022; Khairy et al., 2022), while for others, no information was provided regarding the criteria used for selecting the annotators (Alam et al., 2022; De Smedt et al., 2018; Mohdeb et al., 2022; Obeidat et al., 2022; Raïdy & Harmanani, 2023).

Some few datasets are annotated automatically. For example, Alqmase et al. (2021) annotated automatically the training set of their sport fanaticism dataset using a rule-based approach that relies on a large lexicon of fanatic and unfanatic words. Nakov et al. (2021) annotated, using SVM, a dataset of tweets on misinformation about COVID-19 vaccines. The dataset was then leveraged to analyse the tweets from various perspectives including harmfulness.

Another approach of annotation consists in labelling the examples based on a prior annotation. This approach was utilized, for example, in the dataset described in (Bensalem et al., 2023), wherein The Arabizi examples taken from four existing datasets were merged, and their diverse labels were unified to encompass only *offensive*, and *not offensive*. Another prior-annotation-based technique was employed in (De Smedt et al., 2018). Initially, the authors manually labelled Twitter users as either jihadist or safe, considering various criteria such as keywords, photos, and observed patterns in user names. Subsequently, all tweets extracted from a labelled user were assigned the same label, either *hate* (for tweets extracted from jihadist users) or *safe* (for tweets extracted from non-jihadist users). While this method allows annotating a batch of texts simultaneously, it may lack accuracy, as not all messages written in the timeline of a user spreading toxicity are necessarily toxic.

Finally, if the dataset is composed of synthetic texts, the number of examples within each category could be determined in advance based on the intended characteristics of the dataset to be created. We placed two datasets in this category because both of them are synthetic, although they are quite different. The examples of Röttger's et al. (2022) dataset were not collected from websites or social media platforms but instead created based on predefined patterns of hateful and non-hateful statements. This means that the labels were assigned to the examples at the time of their synthesis. Nonetheless, the created annotation was validated, subsequently, through crowdsourcing. The dataset of Ousidhoum et al. (2021) is particular since it is created for a different task (§4.2.3). The sentences of this dataset are not annotated examples, but instead structured cloze statements that consist of wildcards and causal clauses with masked words. The full sentences can be synthesized by replacing the wildcards with values from predefined lists of social groups, and the masked words are then filled with the large language model. Next, the toxicity of the generated sentences is investigated.

### 4.3.2 Annotator's dialect proficiency

When dealing with multi-dialectal languages, such as Arabic, it is crucial to carefully choose annotators who are native speakers of the specific dialect used in the dataset (Bergman & Diab, 2022). Since Arabic dialect families are significantly different, this consideration ensures that annotators possess a nuanced understanding of the distinct vocabulary, slang, and cultural expressions associated with the dialect of the corpus to be annotated. Most works that mention the annotators' fluency or nativity in specific dialects typically pertain to datasets consisting of a single dialect (*cf.* Table 9). Beyond that, almost 60% of the works in our pool made no mention whatsoever of this critical consideration, while 4 studies provided minimal reporting. Specifically, Alakrot et al. (2018) addressed this issue by selecting annotators whose nationalities were frequently mentioned in the datasets. Mulki and Ghanem (2021a) and Al Bayari and Abdallah (2022) reported that their chosen annotators were proficient in dialects present in the dataset. Additionally, Mubarak, Rashed, et al. (2020) stated that the selected annotator possesses a solid understanding of Arabic dialects in general. On the other hand, many works mentioned that annotators are Arabic native speakers without specifying their dialects.

Table 10. Annotation approaches and number of annotators per text in the Arabic toxic language detection datasets. [+]: One additional annotator is added to validate the annotation and/or to solve the disagreements if any. *: One annotator annotates the whole dataset and 2 or 3 other annotators annotate only a sample to estimate the inter-annotator agreement. [−]: The annotator with the lowest inter-annotation agreement is excluded.

| Annotation approach | | Specific details or #annotators/text when applicable | References |
|---|---|---|---|
| Automatic | | Rule-based classification | (Alqmase et al., 2021, Training set) (Abadji et al., 2022) (Alduailaj & Belghith, 2023) |
| | | Machine-learning based classification | (Nakov et al., 2021) (Jansen et al., 2022) |
| Manual | Crowdsourcing | 3 | (Mubarak et al., 2017) (Albadi et al., 2018) (Alam, Shaar, et al., 2021)[+] (Albadi et al., 2022) (Mubarak, Hassan, et al., 2022)[+] (Röttger et al., 2022) |
| | | 2 – 5 | (Chowdhury et al., 2020) (Alam, Dalvi, et al., 2021) (Shaar et al., 2021) (Nakov et al., 2022) |
| | | 5 | (Ousidhoum et al., 2019) |
| | | ≥ 5 | (Alhelbawy et al., 2016) |
| | | N/A | (Alshalan & Al-Khalifa, 2020) |
| | Selected annotators | 1 | (Albadi et al., 2019)* (Zampieri et al., 2020)* (Hadj Ameur & Aliane, 2021) (Mubarak et al., 2021)* (Mubarak, Rashed, et al., 2020)* (Obeidat et al., 2022)* |
| | | 2 | (Abainia, 2020)[+] (Alshehri et al., 2020) (Mulki & Ghanem, 2021a)[+] (Boucherit & Abainia, 2022)[+] (Fraiwan, 2022)[+] (Raïdy & Harmanani, 2023) |
| | | 3 | (Abozinadah & Jones, 2017) (Alakrot et al., 2018) (Haddad et al., 2019) (Mulki et al., 2019) (Alsafari et al., 2020) (Aldera et al., 2021)[+] (Mulki & Ghanem, 2021b) (Alam et al., 2022)[+] (Al Bayari & Abdallah, 2022) (Almanea & Poesio, 2022) |
| | | 4[−] | (Shannag et al., 2022) (Ousidhoum et al., 2021) |
| | | N/A | (Jamal et al., 2015) (Alqmase et al., 2021, Test set) (Khairy et al., 2022) (M. Habash & Daqour, 2022) (Mohdeb et al., 2022) |
| Extrapolated from a prior annotation | | Derived from the original annotation of the merged datasets | (Awane et al., 2021) (Badri et al., 2022) (Bensalem et al., 2023) |
| | | Derived from the user profile annotation | (De Smedt et al., 2018) |
| Synthetic based on text patterns | | 1 | (Röttger et al., 2022) (Ousidhoum et al., 2021) |

### 4.3.3 Quality and validation

Out of the works we examined, 21 of them incorporated at least one strategy to ensure the accuracy of the manual annotations. We identified three distinct types of these strategies based on when they were applied: before, during, or after the annotation process.

Concerning the pre-annotation measures, one approach, particularly used when the annotation process relies on crowdsourcing, involves verifying the crowdworkers' accuracy by comparing their annotations to those provided by experts. This is typically implemented in a test phase where the crowdworkers are tasked with annotating a small sample of the dataset, previously annotated by experts. In some works, the crowdworkers are evaluated, without notifying them, by incorporating texts from the pre-annotated sample into each crowdworker task. Less accurate crowdworkers are disqualified based on comparison with the expert labels (Albadi et al., 2018, 2022; Alhelbawy et al., 2016; Chowdhury et al., 2020; Mubarak, Hassan, et al., 2022; Shannag et al., 2022). Another approach is selecting crowdworkers with good reputation scores, which are provided on the crowdsourcing platform (Ousidhoum et al., 2019).

Additional approaches are employed, which are providing annotators with a data sample followed by feedback or discussion on their annotations (Alam et al., 2022; Alsafari et al., 2020), offering training to the annotators (Albadi et al., 2019), and taking measures to ensure the language proficiency of crowdworkers (Albadi et al., 2022; Chowdhury et al., 2020).

During the annotation period, in the work of (M. Habash & Daqour, 2022), communication between the annotators and the dataset creators was enabled through a WhatsApp group to answer questions. In (Al Bayari & Abdallah, 2022; Albadi et al., 2019), the annotators' remarks were considered to improve or update the annotation guideline.

After completing the annotation, for nearly all the datasets annotated by a single annotator, additional annotators are recruited to label a sample of the data and the inter-annotator agreement is computed to check the annotation quality. If the number of annotators is two, an additional one is generally brought in to solve the disagreements (see Table 10). Even with a higher number of annotators, some works, such as (Aldera et al., 2021) opted for a validation phase with an additional annotator. In (Alam et al., 2022), the annotators collaborate more closely with a consolidator to reach a consensus on the final label. Conversely, Shannag et al. (2022), excluded all the labels from the annotator who had the lowest inter-annotation agreement with the rest of the annotators, aiming to ensure a unanimous annotation. Some studies chose to delete the cases where one or more annotators failed to select a label (Alakrot et al., 2018; Albadi et al., 2018; Shannag et al., 2022). Others excluded the cases with different labels (Haddad et al., 2019; Mulki et al., 2019; Mulki & Ghanem, 2021b; Raïdy & Harmanani, 2023).

In the case of automatic annotation, the annotation quality is typically estimated by evaluating the performance of the annotation model on a separate dataset, which is the case, for example, in the work of Nakov et al. (2021).

### 4.3.4 Guideline availability

Vidgen & Derczynski (2020) recommend making the annotation guideline publicly available along with the dataset to enhance transparency about the dataset content and encourage other researchers to reuse it. Only 11 out of the analysed works published their annotation guideline, either as a separate document alongside the dataset, in the annotation platform, or on the project website (see Table 11). Besides, while the complete guideline given to the annotators was not shared, 7 works (Al Bayari & Abdallah, 2022; Albadi et al., 2018, 2019, 2022; Alsafari et al., 2020; Mubarak, Rashed, et al., 2020; Shannag et al., 2022) did provide in the dataset-related paper, an extensive description of the annotation process. This included detailed definitions of the annotation labels with illustrative examples, as well as a description of the specific instructions and cautionary notes given to the annotators, sometimes even in cases of special or challenging scenarios such as dealing with sarcasm. Conversely, certain works (such as (Alakrot et al., 2018; Alduailaj & Belghith, 2023; Badri et al., 2022; Khairy et al., 2022; Mazari & Kheddar, 2023)) either did not mention whether they provided a guideline to the annotators or failed to report adequate details about it. The remaining papers (which represent the majority) included little to moderate information about the guidelines, mostly providing generic definitions of the labels with accompanying examples.

Table 11. Accessibility of the annotation guidelines for toxic language datasets.

| Where the guideline is published | References |
| --- | --- |
| Dataset repository | (Ousidhoum et al., 2019) (Alshalan & Al-Khalifa, 2020) (Chowdhury et al., 2020) (Zampieri et al., 2020) (Röttger et al., 2022) |
| Annotation platform | (Alam, Dalvi, et al., 2021) (Alam, Shaar, et al., 2021) (Shaar et al., 2021) (Alam et al., 2022) (Nakov et al., 2022)[13] |
| Project website | (M. Habash & Daqour, 2022)[14] |

### 4.3.5 Inter-annotator agreement

The inter-annotator agreement (IAA) refers to a calculated coefficient that demonstrates the level of consensus among annotators regarding the labels they assigned to data. This coefficient serves to indicate the reliability of the dataset by assuming that higher agreement between annotators correlates with higher reliability (Artstein & Poesio, 2008). However, since the annotation of toxic language could be highly subjective, a low IAA should not be considered a problem (Vidgen & Derczynski, 2020). Nonetheless, it remains important to report it to be considered when analysing the performance of models.

Out of the 54 works we examined where the dataset annotation is manual, 15 works (29%) did not provide any information about the IAA, which is, for instance, the case in (Abozinadah & Jones, 2017)). In the works that did report it, different measures were used, including Fleiss' Kappa, Cohen's Kappa, Krippendorff's Alpha, percentage of agreement, and Gwet's AC1[15]. Some works (Albadi et al., 2018; Boucherit & Abainia, 2022; Mubarak et al., 2017) presented only the values without specifying the used measure.

The reported IAA values vary between 20% (Ousidhoum et al., 2019) and 100% (Mubarak et al., 2021)[16]. Nonetheless, extracting a clear trend from them is not straightforward due to their diverse interpretations. For instance, some works (e.g., (Abainia, 2020; Alakrot et al., 2018; Chowdhury et al., 2020; Mulki et al., 2019; Mulki & Ghanem, 2021b)) reported multiple IAA values corresponding to each pair of annotators. Conversely, other works provided only sub-averages or a single overall average. In some works this coefficient is calculated per attribute (e.g., in (Mulki & Ghanem, 2021c), two IAA are computed separately for the attributes misogyny identification and the misogyny categorisation) or per class (e.g., Mubarak et al. (2021) reported two values corresponding to the classes *Adult* and *Not adult*). Furthermore, while the majority of works computed the IAA on the entire dataset, there are some works where it was calculated on a sample of the dataset that includes extra annotations for validation, which is the case of (Albadi et al., 2019; Chowdhury et al., 2020; Mubarak, Rashed, et al., 2020).

---

[13] All those works published their guidelines along with the project page in the annotation platform: https://micromappers.qcri.org/, and they are accessible by everyone. See for example the annotation guidelines of the different versions of the COVID-19 dataset (Alam, Dalvi, et al., 2021; Alam, Shaar, et al., 2021; Nakov et al., 2022; Shaar et al., 2021), published in each project page in: https://micromappers.qcri.org/project/category/covid-19/
[14] The authors stated that the website (https://sites.google.com/view/arabicmemeshatedet/home) is created specifically to help the annotators understand the task.
[15] See (Artstein & Poesio, 2008) for details on the different IAA measures.
[16] The IAA of 100% was obtained on the *Adult* class. This perfect value is plausible in a dataset of adult content detection, where it is easy to distinguish between the positive and the negative class.

### 4.3.6 Annotators' support

The financial and emotional support of the toxic language annotators is an ethical responsibility. It is considered among the best practices, which have to be considered when building datasets containing harmful content (Bergman & Diab, 2022; Vidgen et al., 2019; Vidgen & Derczynski, 2020). In this context, the association of computational linguistics designed the Responsible NLP Research checklist[17], which contains questions on human annotators: whether they are paid, whether they were informed about the potential risks (e.g., offensive content), and whether the study reports on the way they are recruited.

Out of the analysed works, only five works (Albadi et al., 2022; Chowdhury et al., 2020; M. Habash & Daqour, 2022; Mubarak, Hassan, et al., 2022; Ousidhoum et al., 2021) explicitly mentioned that the annotators were paid[18], and some of them even provided information about the given wages.

When it comes to emotional support, only 2 works commented on that. The study by Albadi et al. (2022) reported that the annotators were warned about the harmfulness of the content to be annotated and were given the freedom to stop the task at any time. Additionally, Röttger et al. (2022) stated that they used Vidgen's et al. (2019) guideline to monitor and protect the annotators' well-being.

## 4.4 Reusability

The primary aim of sharing datasets should be to ensure that they can be easily reused. Specifically, this involves either reusing the dataset to reproduce the experiments conducted by the original authors or using their published results computed on this dataset as a baseline for comparing new experiments. To assess the reusability of the datasets, we examined them based on three criteria: consistency between the paper and the released dataset, availability of division to training and test subsets and availability of the textual content. Below are further details.

### 4.4.1 Data-Paper consistency

Our examination revealed certain cases where the data statistics described in the research paper do not perfectly match the dataset available online, as observed in (Alduailaj & Belghith, 2023; Alshalan & Al-Khalifa, 2020; Awane et al., 2021; Fraiwan, 2022). In (Fraiwan, 2022), for example, the published dataset contains more than 200 additional examples compared to the dataset used in experiments as described in the paper. An extreme case is the work of (Awane et al., 2021), where there is no correspondence between the dataset and the paper, not only in the number of examples but also in the annotation labels and the ratio of offensive examples.

While in the majority of these works, the discrepancies are generally minor, these inconsistencies between the paper and the shared dataset render the results of the authors' experiments impractical for comparison with future studies. It may also lead to unfair results comparison if the authors of future works are not aware of these differences. To prevent this situation, it is essential to explicitly document these variations in the dataset repository and, whenever possible, provide all versions of the dataset.

---

[17] See section D in "Guidelines for Answering Checklist Questions". ACL Rolling Review. Available at https://aclrollingreview.org/responsibleNLPresearch/.
[18] While it can be implied that annotators were paid if the dataset was annotated via a crowdsourcing platform, the listed references include only those works that explicitly disclosed payment information.

### 4.4.2 Training-test split

Many datasets (precisely, 27 datasets) are not released with a division into training and testing sets. Unless authors' experiments are based on cross-validation using the entire dataset, this absence of official splits can hinder others from replicating the authors' results or considering them as a reference for comparison.

### 4.4.3 Textual content availability

Some researchers may opt to include in the dataset post IDs instead of the actual text due to various considerations, often centred around respecting social media platforms' policies (most notably Twitter). This is the case of 8 datasets in our collection (Alam, Shaar, et al., 2021; Albadi et al., 2018, 2019; Alhelbawy et al., 2016; Alsafari et al., 2020; Alshalan & Al-Khalifa, 2020; Jamal et al., 2015; Nakov et al., 2021). When a dataset only contains post IDs, it becomes vulnerable to data loss due to post deletion. This data loss can significantly impact the completeness and the reusability of the dataset, potentially rendering research findings unreproducible.

### 4.4.4 Raw annotations

In addition to the aggregated annotation (a.k.a., gold labels), it would be beneficial to include in the dataset the annotator's raw labels. This broadens the reusability cases of the dataset by enabling the training of classifiers using approaches to learning from disagreement, which do not rely on the assumption that each example has a single gold judgment (Uma et al., 2021).

Within our collection of datasets, there are only 6 datasets that include the raw labels from all annotators (Alakrot et al., 2018; Alhelbawy et al., 2016; Almanea & Poesio, 2022; Mubarak et al., 2017; Mubarak, Rashed, et al., 2020; Röttger et al., 2022).

The misogyny detection dataset, developed by Almanea and Poesio (2022), distinguishes itself as it was specifically designed to investigate annotation disagreements among annotators with diverse gender and belief backgrounds. This dataset was released in the context of the SemEval-2023 learning with disagreements shared task (Leonardelli et al., 2023). In addition to the hard labels generated through a majority vote scheme, the dataset associates with each example the raw labels from all annotators and the soft labels computed from them. These soft labels represent the probability distribution for both *misogyny* and *not misogyny* classes. For instance, if one label indicates *misogyny* and two labels indicate *not misogyny*, the corresponding soft labels for *misogyny* and *not misogyny* are 0.33 and 0.67, respectively. The authors employed two approaches to train a classification model based on the raw annotations: a soft loss training approach treating the task as a multi-label classification and a multi-classifier approach.

## 5 Findings and discussion

This paper represents the first effort to conduct a comprehensive survey of datasets addressing online toxicity specifically within the Arabic language. Below, we provide our findings in response to the four central research questions that guided this investigation.

**RQ1- What are the available Arabic datasets dealing with toxic language?**

Unlike the previous surveys, the wide range of terms we used to search for papers and datasets on the Arabic toxic language allowed us to collect around 54 available datasets published between 2015 and 2023. Only one of them is accessible via payment (Aldera et al., 2021), and the rest are freely available, most notably in public repositories like GitHub. Most of them are collected from Twitter, while other well-known sources such as Wikipedia (Karan & Šnajder, 2019) have not been exploited. These datasets cover all the Arabic dialect families (Egyptian, Maghrebi, Levantine, and Gulf), with the majority of them incorporating multiple dialects. Remarkably, the Algerian dialect is predominant in terms of the number of mono-dialect datasets.

Moreover, Arabizi is scarcely present in toxic language datasets, except for those in the Algerian dialect. Considering the prevalence of this script in social media (Alghamdi & Petraki, 2018; Darwish, 2014; Haghegh, 2021) and the potential variation in offensive language across different dialects (Bensalem et al., 2023), creating additional Arabizi datasets for dialects beyond the currently available ones is essential. This need arises from the fact that offensive words may differ among dialects, posing a challenge to the transfer of knowledge, from one dialect to another, when training classification models (Bensalem et al., 2023).

**RQ2- What are the tasks for which the collected datasets have been created?**

These Arabic datasets have been created to serve different tasks ranging from general toxic language detection and categorisation (40% of the datasets are in this category) to the identification of specific kinds of toxic language such as extremism and misogyny. Toxic language has been also labelled in datasets created for other tasks such as misinformation detection and sentiment analysis. Only two datasets deal with modalities other than text, namely memes (M. Habash & Daqour, 2022) and videos (Albadi et al., 2022), and a single dataset (Ousidhoum et al., 2021) serves as prompts to large language models to assess the toxicity in the machine-generated text (§4.2.3). Nonetheless, it is important to note that the range of tasks addressed in these datasets is still relatively restricted when compared to the diversity found in datasets in English. Below, we outline some aspects that are currently absent or limited in Arabic datasets.

**Subtle or adversarial toxic language** Subtle toxic language conveys harmful meanings without explicit swearing (Hartvigsen et al., 2022). A related concept is adversarial toxic language, wherein the text is intentionally encoded to deceive detection algorithms, such as by altering the spelling of offensive words (Hosseini et al., 2017). Despite the significant challenge in detecting these forms of toxic language, there are currently only a few Arabic datasets addressing this issue. The Multilingual HateCheck dataset (Röttger et al., 2022) involves tagged examples of *implicit derogatory* and *hate speech with spelling variations*. In addition, the misogyny detection datasets (Mulki & Ghanem, 2021a, 2021b) incorporate examples labelled as *stereotyping*, which could be considered a form of implicit toxic language. However, the number of examples in those classes is limited. Therefore, there is a need for a larger Arabic dataset specifically focusing on this topic.

**Explainability** The importance of explainability in toxic language detection systems lies in enabling users to understand why a given text was marked as toxic. On the other hand, it helps stakeholders (e.g., moderators or developers) to identify potential biases or errors in the system's judgments. While this important aspect has been addressed within some English datasets (Mathew et al., 2021; Pavlopoulos et al., 2021), there is currently no toxic language Arabic dataset containing annotations at the span level to enable explainability.

**Counterspeech** In the context of polarized discussions, the constructive and positive remarks (a.k.a., counterspeech) not only serve as exemplary instances to follow, encouraging others to engage in debates (Reimer et al., 2021) but also act as a formidable countermeasure against hate speech (Garland et al., 2022; Siegel & Badaan, 2020; Stark et al., 2020) and demonstrate solidarity with the targeted groups. Hence, to fight against toxic language, in addition to training systems for identifying it, it is essential to train them to recognize counterspeech comments (Garland et al., 2020), making it possible to highlight this positive speech for users. Thus, creating datasets that contain annotated counterspeech and its categories is required to combat toxic language, particularly hate speech, and understand its dynamics (Mathew et al., 2019). Although many datasets in our collection annotate normal or neutral texts, it remains uncertain whether these texts convey positivity and are directed at the victims of toxicity or merely represent irrelevant comments. An exception can be found in the dataset of Mohdeb et al. (2022), where sympathetic comments towards African refugees and migrants are annotated and constitute 21%, and the Multilingual HateCheck functional test (Röttger et al., 2022).

**Degree of toxicity** Some toxicity aspects, most notably those pointing out its severity or its degree (e.g., strong, weak, …etc) (Kivlichan et al., 2021; Sanguinetti et al., 2019; Sharma et al., 2018) are not yet addressed in the Arabic datasets. Identifying such aspects of toxicity is important to allow taking the right action towards it, which may range from just hiding or removing the comment on the social media platform to legal actions. It is also helpful in the context of social and communication studies.

**RQ3- How were these datasets annotated, and were the best annotation practices applied?**

While the majority of the studies offered sufficient information about aspects like the method of dataset annotation (whether manual, crowdsourcing, or automatic), the number of annotators per text, and the inter-annotator agreement, only a small number of them provided access to the guidelines given to the annotators. Moreover, most of the dataset-related papers included only minimal information about the annotation instructions.

Furthermore, our analysis revealed that over half of the works did not report about measures taken to ensure annotation quality, such as assessing annotators' accuracy or language proficiency before or during the annotation task. Besides, the proficiency of annotators in handling the dialects present in multi-dialect datasets is a concern that has not received the attention it deserves. Finally, it is worth noting that the least reported practice in the examined works concerned the well-being and financial support of the annotators.

**RQ4- Can these datasets be easily reused for future research?**

As shown in Section 4.4, it is worth noting that over half of the datasets lack an official division into training and testing subsets. Additionally, some datasets have been released in versions that differ from those described in the published papers. While this does not necessarily hinder researchers from utilizing them, it does not allow considering the results obtained by the dataset creators, on these datasets, as a baseline for future comparisons. Another factor that can render a dataset entirely unusable is the practice of solely publishing social media post IDs instead of the actual text, which is the case for 8 datasets within our collection.

## 5.1 Recommendations and future work

Drawing from both the findings of this survey and the insights assembled from previous works, we recommend that future research should focus on the following areas:

- Leveraging existing datasets to establish benchmarks. It is noteworthy that a few datasets have already been integrated into benchmarking efforts like ORCA (Elmadany et al., 2023), ALUE (Seelawi et al., 2021) and LAraBench (Abdelali et al., 2023). However, the vast majority of the datasets have been exploited in experiments conducted solely by their authors.
- Developing new datasets that focus on challenges not yet addressed in existing ones (such as explainability, counterspeech detection, subtle and adversarial toxic language detection, and the estimation of toxic language severity) and that are collected from sources different from Twitter.
- Striving to implement best practices when constructing and sharing datasets to enhance their reusability and facilitate their fair exploration for model building and evaluation. These practices include establishing detailed annotation guidelines and making them available within the dataset files or in the paper appendix (Klie et al., 2023), sharing the dataset URL in the related publication, and managing the different versions of the dataset, making it clear which version is used in the reported experiments.
- Incorporating ethical considerations into the dataset annotation process by focusing on the well-being of the annotators.

Moreover, a potential avenue for future research in the context of dataset analysis would involve examining datasets for potential biases. Biases within datasets (e.g., topical, temporal, toxic language targets, etc.) pose a significant concern as they diminish the generalizability of models and may increase the risk of developing discriminatory models against certain social categories. Our awareness is limited to a single study (Wich et al., 2022) that reviewed biases in Arabic toxic language datasets. Specifically, the study covered six Arabic datasets and five in English. This accentuates the need for further efforts to analyse the remaining Arabic datasets, which represent a valuable continuation of this work.

# 6    Conclusion

In this survey, we systematically gathered 54 available Arabic datasets related to toxic language detection, along with their corresponding papers. We conducted a thorough analysis of the collected works based on 18 criteria categorized into 4 axes. Through this systematic investigation, we have shed light on the content of existing datasets, offering insights into their annotation process and potential for reuse. The findings revealed issues in the annotation and sharing processes of some datasets, and discerned unexplored areas in the Arabic language, enabling us to formulate recommendations for future works.

Furthermore, the analysed works are listed in a dedicated GitHub repository, creating a centralized resource accessible to researchers and practitioners. This not only enhances the accessibility and visibility of the existing datasets but also facilitates the process of acquiring data for training and testing toxic language detection solutions.

To sum up, we anticipate that this survey will empower researchers and practitioners to make well-informed decisions regarding whether there is a necessity to construct new Arabic toxic language datasets. This will aid in avoiding redundant efforts in developing datasets for tasks that have already been extensively explored. Additionally, by making the datasets accessible in one repository, we hope to assist in data selection decisions, not only for training and testing Arabic toxic language detection solutions but also for data augmentation and generalizability evaluations.

**Acknowledgements**

This publication was made possible by NPRP grant 13S-0206-200281 from the Qatar National Research Fund (a member of Qatar Foundation). The findings achieved herein are solely the responsibility of the authors.

# Appendix

## A     Works obtained outside the systematic search

Table A1 presents details on the acquisition of certain works outside the systematic search. Apart from the dataset from (Hadj Ameur & Aliane, 2021), acquired during our participation in a shared task (de Paula et al., 2023), the other three works consist of papers or datasets that were not publicly available at the time of acquisition. Despite their limited accessibility, we proactively requested these works, motivated by their unique characteristics. We are grateful to the first authors of these works for their response to our requests.

Table A1. Works Acquired Outside Systematic Search.

| Work reference | Acquired work (not available online at the time of acquisition) | How acquired | Comment |
|---|---|---|---|
| (Mohdeb et al., 2022) | Dataset | Contacting the first author | Currently, it is the only work on the detection of anti-refugee hate speech in Arabic. |
| (Raïdy & Harmanani, 2023) | Paper (Now available, obtained before its publication) | Contacting the first author | Currently, it is the only work comprising Arabizi toxic language in the Lebanese dialect. |
| (M. Habash & Daqour, 2022) | Master thesis | Contacting the first author | Currently, it is the only work on the hateful memes detection in Arabic. |
| (Hadj Ameur & Aliane, 2021) | Dataset (Also available upon request) | Participation in CERIST NLP Challenge 2023 | Organizers provided the dataset upon registration in the shared task. The dataset is also described on a dedicated GitHub repository and can be obtained upon request. |

## B     Shared tasks dealing with toxic language detection in Arabic

Table B1 provides details on the datasets used as benchmarks in shared tasks. As previously mentioned (§4, 1$^{st}$ paragraph) certain datasets are first introduced in research papers, and later, other versions of these datasets are utilized in shared tasks and referenced in the overview papers of those tasks.

Table B1. Shared tasks dealing with toxic language detection in Arabic.

| Shared task name (with a hyperlink to its website) | Details | Lang. | Arabic Dataset reference | Overview paper |
|---|---|---|---|---|
| SemEval-2020 Task 12: Multilingual Offensive Language Identification in Social Media (OffensEval 2020) | •Subtask A-Offensive language detection (Binary) | Arabic Danish English Greek | (Mubarak, Rashed, et al., 2020) | (Zampieri et al., 2020) |
| Shared Task on Offensive Language Detection @OSACT 2020 Workshop | •Subtask A- Offensive language detection (Binary).<br>•Subtask B- Hate speech detection (Binary) | Arabic | (Mubarak, Rashed, et al., 2020) | (Mubarak, Darwish, et al., 2020) |
| The first Arabic Misogyny Identification shared task ArMI 2021 a sub-track of HASOC @FIRE2021 | •Subtask A: Misogyny Content Identification (Binary).<br>•Subtask B: Misogyny Behaviour Identification (Multi-class) | Arabic | (Mulki & Ghanem, 2021a) | (Mulki & Ghanem, 2021a)<br>(Mulki & Ghanem, 2021c) |
| NLP4IF-2021 Shared Task on Fighting the COVID-19 Infodemic. | The task comprises, among others, 2 questions on harmfulness detection.<br><br>•Q4-Harmfulness: To what extent is the tweet harmful to the society/person(s)/company(s)/product(s)? (Binary)<br>•Q6- Harmful to Society: Is the tweet harmful to society? (Binary) | Arabic Bulgarian English | (Alam, Dalvi, et al., 2021)<br>(Alam, Shaar, et al., 2021) | (Shaar et al., 2021) |

| | | | | | |
|---|---|---|---|---|---|
| CLEF-2022 CheckThat! Lab Task 1 on Identifying Relevant Claims in Tweets. | •Subtask 1C: Harmful tweet detection. Given a tweet, predict whether it is harmful to society (Binary). | Arabic Bulgarian Dutch English Turkish | (Alam, Shaar, et al., 2021) (Alam, Shaar, et al., 2021) | (Nakov et al., 2022) | |
| Arabic Hate Speech Shared Task 2022@OSACT 2022 Workshop | •Subtask A: Offensive language detection (Binary) •Subtask B: Hate speech detection (Binary) •Subtask C: Detect the fine-grained type of hate speech (Multi-class) | Arabic | (Mubarak, Hassan, et al., 2022) | (Mubarak, Al-Khalifa, et al., 2022) | |
| Overview of the WANLP 2022 Shared Task on Propaganda Detection in Arabic | •Subtask 1: Propaganda techniques identification (multi-class and multi-label) •Subtask 2: Propaganda techniques identification together with the span(s) of text in which each propaganda technique appears (Sequence tagging) | Arabic | (Alam et al., 2022) | (Alam et al., 2022) | |
| CERIST Natural Language Processing (NLP) Challenge 2023. Task 1: Opinion Mining and Sentiment Analysis | Subtask 1.d: Arabic Hate Speech Detection (Binary) | Arabic | (Hadj Ameur & Aliane, 2021) | We are not aware of the overview paper of the task. See instead one of the working note papers (de Paula et al., 2023). The rest of the working notes can be found in this link. | |
| SemEval2023 Task 11 on Learning with Disagreements (LeWiDi) | The task is to use methods for capturing agreements/disagreements in subjective tasks. In Arabic, the task is misogyny identification (Binary). | Arabic English | (Almanea & Poesio, 2022) | (Leonardelli et al., 2023) | |

## C    Datasets list

Table C1 provides the list of the datasets we analysed in this review. See the dedicated repository in GitHub (see endnote 1) for updates.

Table C1. List of the collected datasets.

| Reference | Dataset URL | Size (#examples) | Source |
|---|---|---|---|
| (Jamal et al., 2015) | https://doi.org/10.7910/DVN/28171 | 6077 | Twitter |
| (Alhelbawy et al., 2016) | https://github.com/Alhelbawy/Arabic-Violence-Twitter | 20151 | Twitter |
| (Mubarak et al., 2017) | https://alt.qcri.org/~hmubarak/offensive/AJCommentsClassification-CF.xlsx | 31692 | aljazeera.net |
| (Mubarak et al., 2017) | https://alt.qcri.org/~hmubarak/offensive/TweetClassification-Summary.xlsx | 1100 | Twitter |
| (Abozinadah & Jones, 2017) | https://github.com/eabozinadah/Twitter_Arabic_Abusive_Accounts/blob/master/Abusive%20Accounts%20Database-For%20sharing.zip | 48574 | Twitter |
| (De Smedt et al., 2018) | The original: https://gist.github.com/tom-de-smedt/2d76d33f2515c5a52225af2bb4bb3900  Split to training and testing by another researcher: https://www.kaggle.com/datasets/haithemhermessi/terrorism-and-jihadist-speech-detection | 45000 | Twitter |
| (Albadi et al., 2018) | https://github.com/nuhaalbadi/Arabic_hatespeech | 6136 | Twitter |
| (Alakrot et al., 2018) | https://onedrive.live.com/?authkey=%21ACDXj%5FZNcZPqzy0&id=6EF6951FBF8217F9%21105&cid=6EF6951FBF8217F9 | 15050 | YouTube |
| (Haddad et al., 2019) | https://github.com/Hala-Mulki/T-HSAB-A-Tunisian-Hate-Speech-and-Abusive-Dataset | 6024 | Facebook, YouTube |
| (Mulki et al., 2019) | https://github.com/Hala-Mulki/L-HSAB-First-Arabic-Levantine-HateSpeech-Dataset | 5846 | Twitter |
| (Ousidhoum et al., 2019) | https://github.com/HKUST-KnowComp/MLMA_hate_speech | 3353 | Twitter |
| (Albadi et al., 2019) | https://github.com/nuhaalbadi/ArabicBots | 450 | Twitter |
| (Alshalan & Al-Khalifa, 2020) | https://github.com/raghadsh/Arabic-Hate-speech https://github.com/Ragsh/ArabicHateSpeech | 9316 | Twitter |
| (Mubarak, Rashed, et al., 2020) | https://alt.qcri.org/resources1/OSACT2020-sharedTask-CodaLab-Train-Dev-Test.zip https://edinburghnlp.inf.ed.ac.uk/workshops/OSACT4/datasets | 10000 | Twitter |

| Reference | URL | Size | Source |
|---|---|---|---|
| (Zampieri et al., 2020) | https://drive.google.com/file/d/1GpsLoAPB8MEJvJxeft10tHc643WSrtOW/view | 10000 | Twitter |
| (Alshehri et al., 2020) | https://github.com/UBC-NLP/Arabic-Dangerous-Dataset https://github.com/Nagoudi/Arabic-Dangerous-Dataset | 5011 | Twitter |
| (Chowdhury et al., 2020) | https://github.com/shammur/Arabic-Offensive-Multi-Platform-SocialMedia-Comment-Dataset | 4000 | Twitter, Facebook, YouTube |
| (Alam, Shaar, et al., 2021) | https://github.com/firojalam/COVID-19-disinformation | 4966 | Twitter |
| (Alsafari et al., 2020) | https://github.com/sbalsefri/ArabicHateSpeechDataset | 5341 | Twitter |
| (Abainia, 2020) | https://github.com/xprogramer/DZDC12 | 2400 | Facebook |
| (Mubarak et al., 2021) | https://alt.qcri.org/resources1/AdultContentDetection.zip | 50000 | Twitter |
| (Nakov et al., 2021) | https://gitlab.com/sshaar/a-second-pandemic.-analysis-of-fake-news-about-covid-19-vaccines-in-qatar/-/tree/main/data | 606 | Twitter |
| (Hadj Ameur & Aliane, 2021) | https://github.com/MohamedHadjAmeur/AraCOVID19-MFH | 10828 | Twitter |
| (Mulki & Ghanem, 2021b) | https://github.com/bilalghanem/let-mi | 6550 | Twitter |
| (Mulki & Ghanem, 2021a) | https://github.com/bilalghanem/armi | 9833 | Twitter |
| (Ousidhoum et al., 2021) | https://github.com/HKUST-KnowComp/Probing_toxicity_in_PTLMs | 160552 | Synthetic |
| (Alam, Dalvi, et al., 2021) | http://doi.org/10.7910/DVN/XYK2UE | 218 | Twitter |
| (Shaar et al., 2021) | https://gitlab.com/NLP4IF/nlp4if-2021/-/tree/master/data | 4056 | Twitter |
| (Alqmase et al., 2021) | https://github.com/qumasi/Anti-fanatic-resources/ | 241755 | Twitter |
| (Aldera et al., 2021) | https://ieee-dataport.org/documents/annotated-arabic-extremism-tweets | 89816 | Twitter |
| (Obeidat et al., 2022) | https://doi.org/10.7717/peerj-cs.1151/supp-1 | 6682 | Twitter |
| (Alam et al., 2022) | https://gitlab.com/arabic-nlp/propaganda-detection/-/tree/main/data | 930 | Twitter |
| (M. Habash & Daqour, 2022) | https://github.com/mohamadhabash/Graduation-Project | 1674 | Facebook, Twitter, Reddit |
| (Badri et al., 2022) | https://github.com/NabilBADRI/Multidialect-Project | 23033 | Facbook, Twitter, YouTube |
| (Mubarak, Hassan, et al., 2022) | https://codalab.lisn.upsaclay.fr/competitions/2332#learn_the_details-get_starting_kit | 12698 | Twitter |
| (Boucherit & Abainia, 2022) | https://github.com/xprogramer/DziriOFN | 8749 | Facebook |
| (Nakov et al., 2022) | https://gitlab.com/checkthat_lab/clef2022-checkthat-lab/ | 6155 | Twitter |
| (Almanea & Poesio, 2022) | https://github.com/Le-Wi-Di/le-wi-di.github.io/blob/main/data_post-competition.zip https://codalab.lisn.upsaclay.fr/competitions/6146 | 943 | Twitter |
| (Fraiwan, 2022) | https://data.mendeley.com/v1/datasets/8kftmw7rct/draft?a=0fd4d8fc-42e8-478b-bd17-ba61996aad61 | 24078 | Twitter |
| (Röttger et al., 2022) | https://github.com/rewire-online/multilingual-hatecheck | 3570 | Hand-crafted |
| (Khairy et al., 2022) | https://github.com/omammar167/Arabic-Abusive-Datasets/blob/main/Cyber_2021.csv | 13231 | Twitter, Facebook |
| (Al Bayari & Abdallah, 2022) | https://docs.google.com/spreadsheets/d/1jdG9w3lqCUHilVWk7uVnJhIW1KGOft-K/edit#gid=1549689368 | 47225 | Instagram |
| (Shannag et al., 2022) | https://data.mendeley.com/v1/datasets/z2dfgrzx47/draft?a=12a9ff5d-6c5c-4b2e-8990-7d044d7c12e2 | 4505 | Twitter |
| (Albadi et al., 2022) | https://osf.io/cf9w8/?view_only=aa81f43ff28c4faaa7514ccccc6a386c | 3000 | YouTube |
| (Mohdeb et al., 2022) | Sent by authors upon request. | 4586 | YouTube |
| (Alduailaj & Belghith, 2023) | https://www.kaggle.com/datasets/alanoudaldealij/arabic-cyberbullying-tweets | 26000 | Twitter, YouTube |

| Reference | URL | Size | Source |
|---|---|---|---|
| (Raïdy & Harmanani, 2023) | https://www.kaggle.com/datasets/mariajmraidy/datasets-for-sentiment-analysis-of-arabizi | 3134 | Twitter |
| (Abadji et al., 2022) | https://oscar-project.github.io/documentation/versions/oscar-2201/ | Arabic 8718929 Egyptian Arabic 1256 | Common Crawl |
| (Jansen et al., 2022) | https://oscar-project.github.io/documentation/versions/oscar-2301/ | Arabic 25012116 Egyptian Arabic 3958 | Common Crawl |
| (Awane et al., 2021) | https://drive.google.com/file/d/1MCXY5eyI7myKyQQ2ZPpI1RHPZR7Emd2e/edit | 38654 | Twitter, Facebook, YouTube |
| (Riabi et al., 2023) | https://gitlab.inria.fr/ariabi/release-narabizi-treebank | 1287 | Algerian newspaper's web forums, Song Lyrics |
| (Mazari & Kheddar, 2023) | https://sourceforge.net/projects/alg-dialect-toxicity-speech/ | 14150 | Facebook, YouTube, Twitter |
| (Essefar et al., 2023) | https://github.com/kabilessefar/OMCD-Offensive-Moroccan-Comments-Dataset | 8024 | YouTube |
| (Bensalem et al., 2023) | https://github.com/Imene1/Arabizi-offensive-language | 7383 | Facebook, YouTube, Twitter |